\documentclass[10pt,twocolumn,letterpaper]{article}

\usepackage{cvpr}              % To produce the CAMERA-READY version
\usepackage{array}
\usepackage{multirow}
\usepackage{pifont}
\definecolor{cvprblue}{rgb}{0.21,0.49,0.74}
\usepackage[pagebackref,breaklinks,colorlinks,allcolors=cvprblue]{hyperref}

\def\paperID{18} % *** Enter the Paper ID here
\def\confName{CVPR}
\def\confYear{2026}

\title{Streaming4D: Accelerate 4D World Models via Block-wise Video Generation \\ and Incremental Reconstruction}

\author{
    Xiaoyan Liu$^{1\dagger}$,
    Kangrui Li$^{2\dagger}$,
    Jiaxin Liu$^{3\dagger}$,
    Sifan Zhou$^{4*}$\\
    $^1$The Chinese University of Hong Kong \quad
    $^2$The Hong Kong Polytechnic University \quad \\
    $^3$The University of New South Wales \quad
    $^4$Southeast University \\
    \footnotesize{\texttt{liuxy185@link.cuhk.edu.hk, 24120659G@connect.polyu.hk,
    z5565763@ad.unsw.edu.au, sifanjay@gmail.com}}
}

\begin{document}
\maketitle
\footnotetext[1]{$\dagger$Equal contribution.}
\footnotetext[2]{$*$Corresponding author.}
\begin{abstract}
% Traditional methods for dynamic scene reconstruction are often limited to specific scenarios or require complex multi-stage optimization, making it difficult to efficiently generate generalized 4D scene models with spatiotemporal consistency. To address these challenges, we present a novel cyclic framework that tightly couples autoregressive video generation with 4D reconstruction. Our approach creates a cooperative optimization framework where an initial video sequence, synthesized by an autoregressive model, is lifted into a 4D representation. This representation, in turn, provides geometric feedback to guide the subsequent frame synthesis toward improved consistency. This cyclic guidance mechanism effectively reduces visual drift and ensures robust spatiotemporal coherence across long sequences. Extensive evaluations on large-scale dynamic scene datasets show that our framework achieves high accuracy in 4D reconstruction and produces videos with superior visual fidelity, while maintaining strong generalization to unseen scenes.

Current 4D generation paradigms are often bottlenecked by a sequential decoupling design: video is generated first, followed by 3D reconstruction, leading to high interaction latency. This limits applications in interactive real-time scenarios. To this end, we propose \textbf{Streaming4D}, a tightly coupled synchronous pipeline that integrates block-wise autoregressive video generation with incremental 3D reconstruction. Unlike traditional frame-by-frame emission and delayed geometry recovery, Streaming4D generates temporal video blocks and immediately triggers reconstruction for each completed block, enabling parallel execution between synthesis and geometric updates. This approach allows the world representation to evolve online with the video stream, reducing feedback latency while preserving geometric fidelity. We instantiate \textbf{Streaming4D} using a Self-Forcing-style autoregressive generator and an incremental reconstruction backend. Experiments show consistent runtime improvements across resolutions on a single RTX 4090 (1.24$\times$ speedup), while maintaining high-quality 4D geometry and multi-view consistency.

% This architecture enables the on-the-fly construction of dynamic 3D scenes synchronized with the advancing video stream. Experimental results demonstrate that \textbf{Streaming4D} significantly reduces the overall system feedback latency, achieving up to 1.24× speedup over a sequential baseline, while maintaining high-quality 4D geometry and multi-view consistency.

% Experimental results demonstrate that \textbf{Streaming4D} significantly reduces the overall system feedback latency while maintaining high-fidelity 4D geometric reconstruction quality.
%

\end{abstract}    
\vspace{-6mm}
\section{Introduction}
\label{sec:intro}

%The transition from static 3D reconstruction to dynamic 4D world modeling represents a pivotal shift toward capturing the continuous temporal evolution of the physical world. However, a critical "efficiency-latency gap" persists in current 4D generation paradigms. Most state-of-the-art systems adopt an offline decoupled workflow, where a full video sequence is synthesized before any geometric reasoning begins. This sequential dependency creates a prohibitive Temporal Synchronization Gap, where the 3D world representation lags significantly behind the visual stream, rendering such models unsuitable for high-stakes interactive applications like autonomous navigation or real-time Mixed Reality (MR)

 The transition from static 3D reconstruction to dynamic 4D world modeling represents a pivotal shift toward capturing the continuous temporal evolution of the physical world~\cite{5}. Building such dynamic world models with low latency is increasingly important for interactive applications in vision, graphics, and embodied AI~\cite{xu20244k4d, yan20244dgaussiansplattingscaleaware, ren2024dreamgaussian4dgenerative4dgaussian, zhao2025resilient, zhao2025tartan, hu2026anyslot, zhang2026adaptive, zhang2026multivariate, jiao2026large, zeng2025janusvln,zeng2025FSDrive, zhao2023benchmark,zhao2024balf,zhao2026advances}. Recent progress in video generation and feed-forward 3D reconstruction have provided promising building blocks, yet paratical 4D systems remain constrained by a critical efficiency gap. Most existing pipelines follow an offline decoupled workflow: \textbf{(1)} generate a full video sequence; \textbf{(2)} reconstruct geometry afterward. This sequential dependency introduces prohibitive latency, preventing true interactive 4D synthesis.

To support real-time interaction, both generation and reconstruction must operate causally and continuously, rather than as isolated stages. Under this requirement, autoregressive (AR) modeling has emerged as the powerful architecture for streaming synthesis, as it naturally decomposes the joint distribution of video frames into conditional generation steps~\cite{xiong2025autoregressivemodelsvisionsurvey}. However, existing work such as Rolling Forcing~\cite{liu2025rollingforcingautoregressivelong} and STream3R~\cite{lan2025stream3r} still treat video generation and 3D reconstruction as independent stages in a linear pipeline, leading to computational redundancy and poor temporal synchronization between the synthesized visual flow and the underlying 3D representation. Consequently, when using AR generators, the reconstruction stage often waits for frame-by-frame output, leading to computational redundancy and lack of tight temporal synchronization between the synthesized visual flow and 3D representation.

To address this challenge, we propose \textbf{Streaming4D}, a novel synchronous streaming pipeline that tightly integrates block-wise autoregressive video generation with incremental 3D reconstruction, enabling near real-time, low-latency 4D reconstruction. Instead of the common frame-by-frame emission, \textbf{Streaming4D} generates video segments in discrete chunks to balance temporal coherence and inference throughput. The core innovation of \textbf{Streaming4D} is its pipelined parallelism mechanism, where the AR generator produces each video block and triggers incremental 3D reconstruction process. By overlapping the computation of generative inference and geometric reasoning, \textbf{Streaming4D} ensures the 3D world evolves alongside the advancing video stream, significantly reducing feedback latency and preserving high geometric fidelity. Extensive experiments demonstrate the significantly reduced latency, achieving a speedup of $1.21 \times$ to $1.24 \times$ on a single RTX 4090, while maintaining high-fidelity 4D geometric quality. Our contributions can be summarized as follows:

\begin{itemize}
    \item  \textbf{Synchronous Architecture.} We introduce a tightly coupled synchronous architecture that unifies AR video generation and incremental 4D reconstruction.
 
    \item \textbf{Block-wise Pipeline Strategy.} We design a block-wise pipeline execution strategy that overlaps generation and reconstruction to reduce end-to-end latency.

    \item \textbf{Performance Validation.} We show that the proposed design consistently improves runtime efficiency while preserving geometric quality and multi-view consistency.
\end{itemize}

%-------------------------------------------------------------------------

%-------------------------------------------------------------------------

\begin{figure*}[t]
    \centering
    \includegraphics[width=0.9\linewidth]{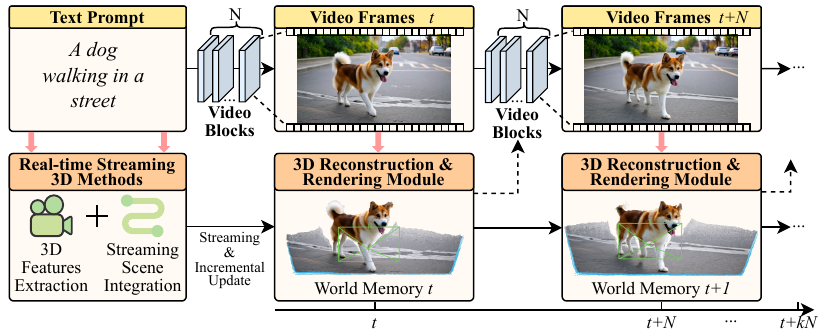}
    \vspace{-4mm}
    \caption{\textbf{Overview of our Streaming4D.} Given a text prompt, our framework enables real-time 4D synthesis via a synchronous, block-wise autoregressive pipeline. It comprises two coupled modules: (1) \textbf{Video Generation}, producing contiguous $N-frame$ video blocks($N = 3m$), and (2) \textbf{3D Reconstruction}, which integrates features into a persistent \textbf{World Memory}. By parallelizing the denoising of block $B_{n+1}$ with the reconstruction of block $B_n$, our system achieves incremental, real-time world modeling. The dotted line denotes an optional geometric consistency constraint.}
    \label{fig:pipline}
    \vspace{-5mm}
\end{figure*}

\vspace{-8mm}
\section{Related work}
\label{sec:rw}
\vspace{-3mm}
\textbf{Autoregressive Video Generation.} Autoregressive models\cite{van2016pixel, villegas2022phenakivariablelengthvideo, VAR, henschel2025streamingt2v} have become the dominant approach for sequential video synthesis.
% excelling at capturing complex dependencies and preserving temporal coherence. 
By factorizing the joint distribution into conditional generation steps, AR models have propelled the shift from batch generation to streaming synthesis for real-time applications\cite{gu2025starflow, liu2025rollingforcingautoregressivelong}. 
To address the challenges of error accumulation and slow inference, 
% To address these issues,
% the Self Forcing~\cite{huang2025selfforcingbridgingtraintest} align training distributions with inference states through a global matching loss, mitigating drift during long-term rollouts. Meanwhile, CausVid~\cite{yin2025slowbidirectionalfastautoregressive} distills bidirectional diffusion into an AR generator, and the NFD framework~\cite{cheng2025playingtransformer30fps} achieves good speeds via intra-frame parallel sampling. 
the Self Forcing~\cite{huang2025selfforcingbridgingtraintest} aligns train-test distributions via global matching to reduce rollout drift, CausVid~\cite{yin2025slowbidirectionalfastautoregressive} distills bidirectional diffusion into AR generation, and NFD~\cite{cheng2025playingtransformer30fps} accelerates inference via intra-frame parallel sampling.
% However, most AR video generation models focus on 2D content and lack the explicit geometric reasoning required for 4D scene construction. 
However, most AR generators focus on 2D content that lacks explicit geometric reasoning.
Recently, AR4D \cite{zhu2025ar4d} extends autoregressive paradigms to 4D generation. Despite sharing a similar motivation for continuous 4D world modeling, our Streaming4D uniquely distinguishes itself by introducing a block-wise pipelined synchronization strategy to minimize interaction latency between video emission and geometry recovery.
 
% The core advantage of autoregressive generation lies in its powerful sequence modeling capability, enabling it to capture complex dependencies and dynamic patterns within data.
% However, generating high-quality, long-term, and coherent videos still faces core challenges, including error accumulation and slow generation speed. To address these issues, the research community has developed a clear technological trajectory, where early explorations in video generation not only provided key solutions for temporal error control and computational efficiency but also established a critical foundation for building complex 4D generation systems. The self-force method~\cite{huang2025selfforcingbridgingtraintest} predicts the next frame based on the history generated by the model itself and employs a global distribution matching loss to reduce the accumulation of errors during inference.
% Meanwhile, CausVid~\cite{yin2025slowbidirectionalfastautoregressive} technology distills a bidirectional diffusion model into an autoregressive generator, achieving video streaming generation. The NFD framework~\cite{cheng2025playingtransformer30fps} achieves breakthrough real-time performance through its innovative combination of parallel sampling in the intra-frame and autoregressive methodology in the inter-frame. These techniques provide crucial insights for realizing real-time, multi-modal (text/image-to-4D) generation within a unified architecture.

% 要不要写以4D重建为先验指导视频生成？

\noindent\textbf{Integrated 4D World Models.} On the reconstruction side, feed-forward 4D generation techniques\cite{chen20254dnex, wang20254real, pan2025diff4splat, liu2025dream4d} alongside 
others\cite{fisher2021colmap, orbslam, 4dgs, kerbl20233d, yin20234dgen, wu2025cat4d} prioritize the reconstruction of geometrically consistent dynamic scenes. Achieving this fidelity is crucial for developing interactive world models, yet existing methods struggle to balance fidelity and real-time responsiveness. 
% which require tight integration of generative priors and geometric reconstruction. However, existing methods often struggle to balance high-fidelity consistency and real-time responsiveness. 
For example, DUSt3R~\cite{wang2024dust3r} focuses on static stereo pairs, and MASt3R~\cite{murai2025mast3r} extends this to video streams, but these methods often require global or window-based optimization, which poses challenges for real-time 4D synthesis. VGGT~\cite{wang2025vggt} and its streaming variant StreamVGGT~\cite{zhuo2025streaming} offer robust zero-shot geometric reasoning but still face sequential offline bottlenecks, delaying reconstruction and introducing prohibitive latency. 
% offer robust zero-shot geometric reasoning but at the cost of batch processing. This reliance create offline bottleneck, delaying reconstruction until a full video sequence is completed, as seen in Stream3R~\cite{zhuo2025streaming}.
% This fundamental reliance on batch or delayed processing contributes to the persistent offline decoupling bottleneck in current 4D synthesis paradigms, where reconstruction can only begin after the entire video sequence is finalized like Stream3r~\cite{zhuo2025streaming}. 
% This sequential pattern introduces prohibitive latency, which makes them unsuitable for real-time interaction.
% In parallel to generative modeling, 4D generation techniques\cite{pumarola2021d, kerbl20233d, yin20234dgen} prioritize the reconstruction of geometrically consistent dynamic scenes.
% Integration of generation and reconstruction is the cornerstone of the development of interactive world models. While DUSt3R~\cite{wang2024dust3r} focuses on static stereo pairs and MASt3R~\cite{murai2025mast3r} extends this to video streams through pointmap tracking, they often require global or window-based optimization, which poses challenges for real-time 4D synthesis. This fundamental reliance on batch or delayed processing contributes to the persistent offline decoupling bottleneck in current 4D synthesis paradigms, where reconstruction can only begin after the entire video sequence is finalized like Stream3r~\cite{zhuo2025streaming}. This sequential pattern introduces prohibitive latency, which makes them unsuitable for real-time interaction. 
Another line of research, represented by Point3R~\cite{wu2025point3r} and other methods based on streaming-memory \cite{sandstrom2023point, cheng2024gaussianpro}, maintains an explicit spatial memory to allow incremental updates of large-scale environments. For example, TeleWorld~\cite{chen2025teleworld} introduces a "Generation-Reconstruction-Guidance" loop to synchronize 4D field updates with video generation. In this paper, we advance the concept via block-wise synchronization, enabling incremental 3D reconstruction for each video block.  
\section{Method}
\label{sec:Method}

\subsection{Block-wise Autoregressive Video Generation}
The first stage of our pipeline is dedicated to generating a continuous video stream from the initial conditions. To balance inference throughput and spatiotemporal consistency, we move beyond conventional frame-by-frame emission in favor of a Block-wise AR strategy.

The generation process begins with a text prompt $\mathcal{P}$ that describes the desired dynamic scene. To achieve seamless streaming output, we decompose the joint distribution of the video sequence into a series of conditional generative steps, where video frames are produced in discrete spatiotemporal blocks $\left\{B_{k}\right\}_{k=1}^{K}$. Each block $B_k$ contains $N$ consecutive frames (e.g., in our implementation, we use $ N = 3m$ frames, where $m \in \mathbb{Z}$.).

% Our goal is to synthesize a sequence of video blocks $B_t = {I_{t,1}, I_{t,2},..., I_{t,N}}$, where $N$ represents the block size (e.g., $N=3$ frames per block in our settings). The generation process is formulated as a conditional denoising process. 

Let $G_{\mathrm{AR}}$ denote the AR video generator. To synthesize the 
$k-$th segment of the video stream, the generator predicts the current video block $B_k$ conditioned on the text prompt embeddings $\tau(\mathcal{P})$ and the history context provided by the previously generated block $B_{k-1}$. Formally, the AR generation is defined as: 
\vspace{-2mm}
\begin{equation}
\begin{aligned}
    B_{k}=G_{\mathrm{AR}}\left(B_{k-1}, \tau(\mathcal{P}), z_{k} ; \Theta_{\text {video }}\right),
\end{aligned}
\end{equation}
\vspace{-3mm}

% For the initial step $t=1$, 
where $z_k$ is the sampled latent noise for the current block, and $\Theta_{\text {video }}$ represents the model parameters. For the initial block ($n=1$), the generation is conditioned solely on the text prompt: 

\vspace{-2mm}
\begin{equation}
\begin{aligned}
B_{1}=G_{\mathrm{AR}}\left( \tau(\mathcal{P}), z_{1} ; \Theta_{\text {video }}\right),
\end{aligned}
\end{equation}
\vspace{-1mm}
To mitigate the error accumulation commonly observed in long-term AR rollouts, our model follows the Self-Forcing paradigm~\cite{huang2025selfforcingbridgingtraintest}, which aligns the training condition distribution with the inference state, ensuring long-term spatiotemporal consistency. Crucially, once the block generation $B_k$ is completed, it is immediately sent to the 3D reconstruction module, while the generator seamlessly proceeds to denoise the next block $B_{k+1}$.

\subsection{Real-time Incremental 4D Reconstruction}
The primary objective of our framework's backend is to facilitate the online incremental construction of dynamic 4D scenes. To achieve this, we introduce a real-time 4D reconstruction pipeline that seamlessly consumes the video blocks $B_n$ emitted by the AR generator.

Instead of waiting for the entire video sequence, our 4D generator updates a persistent 3D state representation on the fly. We maintain a persistent state represented as a set of token embeddings $S$ that evolve sequentially. 
% Upon receiving a newly generated video block $B_t$, the 4D generator processes its constituent frames $I \in B_t$ to perform two key operations simultaneously:

% Our 4D generator maintains a persistent state represented as a set of token embeddings that evolve with each new frame. Specifically, for each generated frame $I_t$, the 4D generator performs two key operations simultaneously.

\noindent
\textbf{State Update}: The persistent state $S_k$ is updated by integrating information from the current observation:
\vspace{-2mm}
\begin{equation}
\begin{aligned}
S_{k}=\textit{Update} \textit{Transformer}\left(S_{k-1}, E\left(B_{k}\right)\right),
\end{aligned}
\end{equation}
% \vspace{-1mm}
where $E(B_k)$ represents the visual features extracted from the video block $B_k$ by a Vision Transformer encoder, and $\operatorname{Update} \operatorname{Transformer}$ is a specialized decoder that merges new observations with the existing state. By retaining and updating historical scene information at each step, state tokens provide a continuous context that allows the model to handle dynamic content consistently.

\noindent
\textbf{State Readout}: Based on updated state, the model predicts metric-scale point maps in camera and world coordinates:
\vspace{-2mm}
\begin{equation}
\begin{aligned}
X_k^{cam},X_k^{world}=ReadoutTransformer(S_k),
\end{aligned}
\end{equation}
% \vspace{-1mm}
where $X_k^{cam}$ and $X_k^{world}$ represent the 3D point clouds in camera and world coordinates, respectively. The world coordinate pointmaps are accumulated over time to form a coherent 4D reconstruction. 

% The combination of AR video generation paradigm and 4D generator creates a powerful framework for maintaining temporal consistency in 4D reconstruction. The 4D generator's persistent state mechanism naturally ensures spatial coherence without explicit alignment procedures. Meanwhile, the AR video generation model addresses the temporal coherence at the frame generation level through its distribution matching objective, which ensures that long-term dynamics remain realistic and coherent. The integration of these two mechanisms provides comprehensive spatio-temporal consistency throughout the 4D reconstruction pipeline.

% The state tokens act as a memory bank that accumulates scene information over time, allowing the model to maintain consistency even with dynamic content. This is formally represented as:

\vspace{-2mm}
\subsection{Pipelined Parallelism and Synchronization}
\label{sec:3.3}
 % Building upon the state tokens that serve as a compact scene representation, we now introduce a comprehensive guidance framework for video generation that ensures both structural fidelity and temporal coherence.  Our approach leverages multiple complementary guidance signals to maintain persistent world memory throughout the video synthesis process.
 
 % The core efficiency of our framework stems from its pipelined parallelism mechanism, which overlaps the generative inference with geometric reasoning. Unlike traditional decoupled 4D synthesis paradigms where 3D reconstruction $R$ must wait for the entire video sequence to be finalized, our system treats video generation and 4D reconstruction as a asynchronous parallel processing framework operating in a staggered temporal window.

 We adopt a block-wise strategy rather than frame-by-frame emission to balance inference throughput and temporal coherence. Each block serves as a compact spatiotemporal unit that captures local motion dynamics more effectively. Crucially, this granularity enables pipelined parallelism, allowing the 3D reconstruction $R$ of block $B_{k-1}$ to overlap with the generation of $B_k$. 
 % This staggered execution effectively hides computational overhead, ensuring real-time responsiveness as the 3D scene evolves alongside the video stream

 Let $\mathcal{T}_{gen}(B_k)$ denote the time required to synthesize a video block $B_n$, and $\mathcal{T}_{rec}(B_k)$ denote the time it takes for the 4D reconstruction module to update the world memory and readout the 3D geometry from $B_n$. In a conventional sequential pipeline, the total latency $\mathcal{L}$ for $k$ blocks is $
\sum_{i=1}^{K} \left( \mathcal{T}_{\text{gen}}(B_i) + \mathcal{T}_{\text{rec}}(B_i) \right).$

 In our synchronous streaming pipeline, the two modules execute in parallel with a single-block offset. While the generator produces the block $B_k$, the reconstruction module simultaneously processes the previously completed block $B_{k-1}$. The execution state $\Psi_k$ at the wall-clock step $k$ can be formulated as:

\vspace{-3mm}
 \begin{equation}
 \begin{aligned}
 \Psi_n = \{ G_{AR}(B_{k-1}, \mathcal{P}) \parallel R_{4D}(B_{k-1}, S_{k-2}) \},
  \end{aligned}
 \end{equation}
 \vspace{-1mm}

 where $\parallel$ denotes the execution of concurrently on the computing device. Crucially, this asynchronous execution mode hides the computational overhead, ensuring that geometric reasoning does not become a bottleneck for streaming synthesis. To quantify the efficiency gain, let $\Delta_i=\max\big(\mathcal{T}_{gen}(B_i), \mathcal{T}_{rec}(B_{i-1})\big)$ denote the effective duration of the $i-$th overlapped processing step. The total optimized latency for generating $K$ blocks can then be formulated as:
 
 % Under this mechanism, defining $\Delta_i=\max\big(\mathcal{T}_{gen}(B_i), \mathcal{T}_{rec}(B_{i-1})\big)$, the cumulative latency for generating $K-$ block sequence is optimized to:

% \vspace{-2mm}
%  \begin{equation}
%  \begin{aligned}
%      \begin{split}
% \mathcal{L}_{ours} &= \mathcal{T}_{gen}(B_1) + \sum_{i=2}^{K} \max\big(\mathcal{T}_{gen}(B_i), \mathcal{T}_{rec}(B_{i-1})\big) \\
% &\quad + \mathcal{T}_{rec}(B_K).
% \end{split}
% \end{aligned}
% \end{equation}

% \vspace{-2mm}
% \begin{equation}
% \begin{aligned}
% \mathcal{L}_{ours} = \mathcal{T}_{gen}(B_1) + \sum_{i=2}^{K} \Delta_i + \mathcal{T}_{rec}(B_K).
% \end{aligned}
% \end{equation}

\vspace{-4mm}
\begin{equation}
\begin{aligned}
\mathcal{L}_{ours} = \mathcal{T}_{gen}(B_1) + \sum_{i=2}^{K} \Delta_i + \mathcal{T}_{rec}(B_K).
\end{aligned}
\end{equation}
\vspace{-2mm}
 
% Since our 4D reconstruction module is highly optimized for incremental updates (e.g., $\mathcal{T}_{rec}(B_i)<\mathcal{T}_{gen}(B_i)$), the system's overall throughput is effectively governed by the video generation speed alone, while the 3D world "evolves" in the background with negligible overhead.

\begin{figure*}[t]
    \centering
    \includegraphics[width=0.9\linewidth]{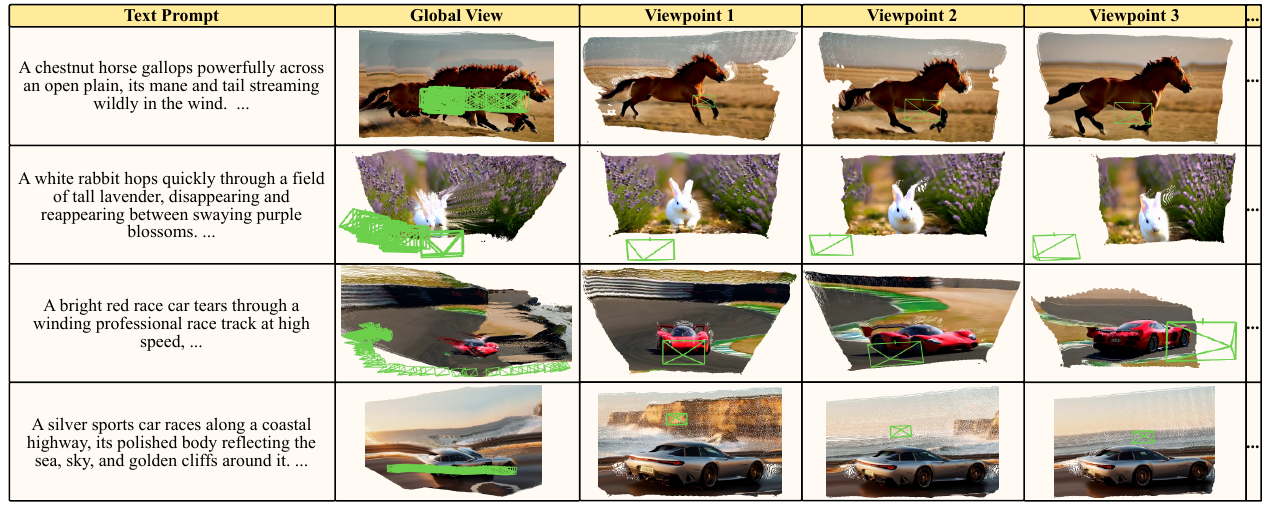}
    \vspace{-2mm}
      \caption{\textbf{Qualitative results of Streaming4D.} Driven purely by text prompts, our block-wise AR generator and continuous 3D reconstructor operate synchronously to model dynamic scenes. The \textbf{Global View} reveals the underlying 3D point cloud structure and the camera poses (green boxes) tracked over time. The \textbf{Viewpoints} illustrate the framework's ability to provide geometrically grounded visual synthesis across different viewing angles and timesteps.}
    \label{quality}
    \vspace{-4mm}
\end{figure*}

While Appendix details hardware contention during actual execution, our current forward-streaming pipeline effectively minimizes interaction latency. Importantly, the modular architecture of Streaming4D inherently supports a closed-loop feedback mechanism. As discussed in Appendix, integrating this geometric guidance will be the focus of our future work to further mitigate cumulative errors in long-duration rollouts.

% The ideal computing process is as described above, while the actual loss can be referenced in \cref{sec:LOHC}. Although our current pipeline primarily focuses on the forward-streaming execution to minimize interaction latency, the modular design of Streaming4D inherently supports a feedback loop. The preliminary analysis in \cref{sec:FW} suggests that this path is the key to mitigating cumulative errors in long-duration rollouts, which will be the focus of our subsequent experiments.

% \vspace{-1mm}
\section{Experiments}
\label{sec:Experiment}

\vspace{-2mm}
% In this section, we introduce the implementation details such as the model architecture, datasets, evaluation metrics, and comparison baselines (see xx Section), as well as the evaluation results included on xx tasks (see xx Section).

% \begin{table}[htbp]
%   \centering
%   \caption{Generation Time Comparison at Different Input Resolutions of CUT3R~\cite{wang2025continuous} on vGPU-32GB (Unit: seconds)}
%   \label{tab:time_comparison}
%   \begin{tabular}{c c c c}
%     \toprule
%     \textbf{Resolution} & \textbf{Baseline} & \textbf{Our Method} & \textbf{Speedup} \\
%     \midrule
%     384$\times$208  & 36.81  & 25.69  & 1.43$\times$ \\
%     512$\times$288  & 41.07 & 33.81  & 1.21$\times$ \\
%     640$\times$368 & 47.62 & 38.33 & 1.24$\times$ \\
%     \bottomrule
%   \end{tabular}
% \end{table}

% \noindent
% \textbf{Architecture}.
% We employ a streaming generation strategy to enable real-time synthesis of 4D content, which is further refined through a mutual-feedback loop that continuously optimizes both video generation and 4D reconstruction processes. 

\textbf{Implementation Detail.} Our framework is implemented as a tightly coupled synchronous pipeline, integrating a Self Forcing-based~\cite{huang2025selfforcingbridgingtraintest} generator for block-wise video synthesis and a CUT3R-based~\cite{wang2025continuous} backend for incremental 4D scene updates (see Appendix for further implementation details).

\begin{table}[htbp]
  \centering
  \vspace{-2mm}
  \caption{Generation time comparison at different resolutions of CUT3R~\cite{wang2025continuous} on single RTX 4090 (Unit: seconds)}
  \vspace{-2mm}
  \label{tab:time_comparison}
  \begin{tabular}{p{2cm} c c c c}
    \toprule
    \textbf{Resolution} & \textbf{Baseline} & \textbf{Our Method} & \textbf{Speedup} \\
    \midrule
    384$\times$208  & 35.64  & 29.29  & 1.21$\times$ \\
    448$\times$256 & 37.09 & 30.04 & 1.23$\times$ \\
    512$\times$288  & 39.29 & 31.61  & 1.24$\times$ \\
    640$\times$368 & 44.13 & 36.33 & 1.21$\times$ \\
    \bottomrule
  \end{tabular}
  \vspace{-4mm}
\end{table}

% For 4D scene generation, we incorporate a module based on the CUT3R~\cite{wang2025continuous} architecture, extended to operate in an online feedback mode. For each generated three-frame block, the 4D generator incrementally updates a persistent world state that encodes evolving geometry, depth, and appearance features, producing a temporally coherent 4D representation. 
% Unlike the original Cut3R, which reconstructs 4D geometry only after the full video is produced, our modified version exposes intermediate latent features and feeds them back to the video generator as high-level structural guidance through cross-attention layers.
% This feedback mechanism allows the 4D generator to provide continuous guidance for preceding video generation, improving spatial consistency, and reducing temporal drift.

% \noindent
% \textbf{Metrics}.

% \noindent
% \textbf{Baseline}.
\vspace{-2.5mm}
\subsection{Quantitative Results}
\vspace{-1mm}
\label{sec:expres}
% \vspace{-2mm}
\begin{table}[htbp]
  \centering
  \caption{\textbf{Comparative 3D reconstruction on the 7-Scenes dataset~\cite{7-scenes}.} On certain metrics, it even achieves performance comparable to CUT3R~\cite{wang2025continuous}.}
  \vspace{-2mm}
  \label{tab:7scenes_results}
  \resizebox{\linewidth}{!}{
    \begin{tabular}{l c c c c c c}
      \toprule
      Method & \multicolumn{2}{c}{Acc$\downarrow$} & \multicolumn{2}{c}{Comp$\downarrow$} & \multicolumn{2}{c}{NC$\uparrow$} \\
      \cmidrule(lr){2-3} \cmidrule(lr){4-5} \cmidrule(lr){6-7}
      & Mean & Med. & Mean & Med. & Mean & Med. \\
      \midrule
      % DUSt3R-GA [107] & 0.146 & 0.077 & 0.181 & 0.067 & 0.736 & 0.839 \\
      % MASt3R-GA [51]  & 0.185 & 0.081 & 0.180 & 0.069 & 0.701 & 0.792 \\
      MonST3R~\cite{zhang2024monst3r}   & 0.248 & 0.185 & 0.266 & 0.167 & 0.672 & 0.759 \\
      Spann3R~\cite{Spann3R}   & 0.298 & 0.226 & 0.205 & 0.112 & 0.650 & 0.730 \\
      CUT3R~\cite{wang2025continuous}   & \underline{0.126} & \underline{0.047} & \textbf{0.154} & \underline{0.031} & \textbf{0.727} & \textbf{0.834} \\
      \textbf{Streaming4D} (Ours)    & \textbf{0.124} & \textbf{0.033} & \underline{0.162} & \textbf{0.027} & \underline{0.705} & \underline{0.810} \\
      \bottomrule
    \end{tabular}
  }
  \vspace{-3.5mm}
\end{table}

% In this section, we report the qualitative and quantitative results of our comprehensive experiments.  

\textbf{Generation Latency}. To evaluate the computational efficiency of our proposed framework, we compare the inference time with the baseline under various input resolutions of CUT3R~\cite{wang2025continuous}. 
% \cref{tab:time_comparison} presents the generation time and speedup ratios of our proposed pipeline compared to the sequential baseline across different input resolutions on a single RTX 4090 GPU. The baseline represents a serial method that simply concatenates the video generation module and the 3D reconstruction module.
\cref{tab:time_comparison} compares the inference latency of our pipeline with a naive sequential baseline on a single RTX 4090 GPU. 
% At lower resolutions (e.g. $384 \times 208$ and $512 \times 288$), our approach reduces the generation time, achieving speedups of $1.21 \times$ and $1.24 \times$, respectively. Furthermore, this efficiency advantage is preserved even at higher resolutions such as $640 \times 368$, where our method maintains a $1.21 \times$ speedup over the baseline, further demonstrating its robust inference speed. 
By overlapping video generation and 3D reconstruction, our approach consistently achieves a speedup $1.21 \times$ to $1.24 \times$ across all resolutions tested (from $384 \times 208$ to $640 \times 368$). This sustained efficiency demonstrates the robustness and scalability of our parallel processing mechanism, even under high-resolution workloads.
The more relevant analysis can be found in Appendix.

% To evaluate the computational efficiency of our proposed framework, we compare the inference time with the baseline under various input resolutions of CUT3R~\cite{wang2025continuous}. As summarized in \cref{tab:time_comparison}, our approach consistently achieves notable acceleration in all evaluated resolutions. At lower resolutions (e.g. $384 \times 208$ and $512 \times 288$), our approach reduces the generation time, achieving speedups of $1.43 \times$ and $1.21 \times$, respectively. Furthermore, this efficiency advantage is preserved even at higher resolutions such as $640 \times 368$, where our method maintains a $1.24 \times$ speedup over the baseline, further demonstrating its robust inference speed. 
% This consistent reduction confirms the practical efficiency of our framework.

% Interestingly, we observe an inverted-U performance curve in relative speedup, which peaks at $1.24\times$ at the $512 \times 288$ resolution, before plateauing or slightly degrading at both lower and higher extremes. 

\noindent
\textbf{3D Reconstruction}. \cref{tab:7scenes_results} presents the quantitative results of our 3D reconstruction quality on the widely used 7-Scenes dataset~\cite{7-scenes}. We evaluate our approach using three standard geometric metrics: Accuracy (Acc), Completeness (Comp), and Normal Consistency (NC). As shown in the table, Our method outperforms conventional approaches (e.g., MonST3R~\cite{zhang2024monst3r}, Spann3R~\cite{Spann3R}) across all metrics. Moreover, compared to the key component CUT3R~\cite{wang2025continuous}, our entire pipeline matches its performance without error accumulation, demonstrating that our framework preserves the quality of the 3D reconstruction.
% our method demonstrates significant superiority over conventional reconstruction approaches such as MonST3R~\cite{zhang2024monst3r} and Spann3R~\cite{Spann3R} for all metrics evaluated. Most importantly, we compare our entire pipeline against CUT3R~\cite{wang2025continuous}, which serves as a fundamental component within our system. Instead of suffering from the error accumulation common in complex pipelines, our method achieves performance comparable to CUT3R. These quantitative results demonstrate that our proposed framework maintains the quality of its underlying 3D reconstruction.

\vspace{-1mm}
\subsection{Qualitative results} 
% As shown in \cref{quality}, we present qualitative results for challenging text prompts. Our method synthesizes high-fidelity, multi-view consistent scenes that adhere closely to textual descriptions. Specifically, the rendered Viewpoints 1-3 showcase the model's robustness in handling dynamic subjects and complex backgrounds. For instance, in the first and second rows, the anatomical structure of the galloping horse and the fine-grained details of the rabbit in the lavender field are consistently preserved across significant viewpoint variations, avoiding common geometric distortions. Crucially, Global View confirms that our model successfully reconstructs a coherent global spatial representation, supporting expansive and complex camera trajectories (indicated by the green frustums) without background collapse or temporal flickering.
Qualitative results in \cref{quality} highlight our framework's ability to generate dynamic high-fidelity, text-aligned scenes. As seen in \textbf{Viewpoints 1-3}, our method robustly handles complex dynamics and backgrounds, preserving fine structural details (e.g., the horse's anatomy) across significant viewpoint shifts while avoiding geometric distortions. Moreover, \textbf{Global View} visualizes a coherent, artifact-free spatial representation that supports expansive camera trajectories (green frustums) without background collapse.

% Beyond quantitative metrics, \cref{quality} demonstrates the robustness of our synchronous generation pipeline on highly dynamic scenes. Fast motions such as a galloping horse or a speeding race car typically cause severe geometric artifacts and temporal flickering in conventional 4D synthesis. However, our method successfully overcomes these challenges by maintaining strict multi-view consistency. The rendered viewpoints reveal that fine anatomical details and complex backgrounds remain stable across extreme camera shifts. Furthermore, the global point cloud confirms that our persistent world memory accurately accumulates spatial information over time. This continuous integration ensures stable camera trajectories and fundamentally prevents background collapse during long-duration generation.

% \input{sec/5_analysis}
\vspace{-2mm}
\section{Conclusion}
\vspace{-2mm}

In this paper, we present \textbf{Streaming4D}, a synchronous framework for low-latency 4D world modeling that tightly couples block-wise autoregressive video generation with incremental 3D reconstruction. By replacing conventional decoupled workflows with block-level pipelined execution, Streaming4D enables online world representation updates, significantly reducing latency without compromising geometric fidelity. Experiments on a single RTX 4090 demonstrate a 1.24$\times$ speedup over sequential baselines while maintaining high temporal stability. These results underscore the potential of integrated pipelines for interactive 4D modeling. Future work will explore closing the loop by feeding reconstructed states back to the generator.

{
    \small
    \bibliographystyle{ieeenat_fullname}
    \bibliography{main}

@inproceedings{5,
  title={Neuralrecon: Real-time coherent 3d reconstruction from monocular video},
  author={Sun, Jiaming and Xie, Yiming and Chen, Linghao and Zhou, Xiaowei and Bao, Hujun},
  booktitle={Proceedings of the IEEE/CVF conference on computer vision and pattern recognition},
  pages={15598--15607},
  year={2021}
}

@article{lan2025stream3r,
  title={Stream3r: Scalable sequential 3d reconstruction with causal transformer},
  author={Lan, Yushi and Luo, Yihang and Hong, Fangzhou and Zhou, Shangchen and Chen, Honghua and Lyu, Zhaoyang and Yang, Shuai and Dai, Bo and Loy, Chen Change and Pan, Xingang},
  journal={arXiv preprint arXiv:2508.10893},
  year={2025}
}

@inproceedings{xu20244k4d,
  title={4k4d: Real-time 4d view synthesis at 4k resolution},
  author={Xu, Zhen and Peng, Sida and Lin, Haotong and He, Guangzhao and Sun, Jiaming and Shen, Yujun and Bao, Hujun and Zhou, Xiaowei},
  booktitle={Proceedings of the IEEE/CVF conference on computer vision and pattern recognition},
  pages={20029--20040},
  year={2024}
}

@inproceedings{wang2024dust3r,
  title={Dust3r: Geometric 3d vision made easy},
  author={Wang, Shuzhe and Leroy, Vincent and Cabon, Yohann and Chidlovskii, Boris and Revaud, Jerome},
  booktitle={Proceedings of the IEEE/CVF conference on computer vision and pattern recognition},
  pages={20697--20709},
  year={2024}
}

@inproceedings{murai2025mast3r,
  title={Mast3r-slam: Real-time dense slam with 3d reconstruction priors},
  author={Murai, Riku and Dexheimer, Eric and Davison, Andrew J},
  booktitle={Proceedings of the Computer Vision and Pattern Recognition Conference},
  pages={16695--16705},
  year={2025}
}

@misc{yan20244dgaussiansplattingscaleaware,
      title={4D Gaussian Splatting with Scale-aware Residual Field and Adaptive Optimization for Real-time Rendering of Temporally Complex Dynamic Scenes}, 
      author={Jinbo Yan and Rui Peng and Luyang Tang and Ronggang Wang},
      year={2024},
      eprint={2412.06299},
      archivePrefix={arXiv},
      primaryClass={cs.CV},
      url={https://arxiv.org/abs/2412.06299}, 
}

@misc{ren2024dreamgaussian4dgenerative4dgaussian,
      title={DreamGaussian4D: Generative 4D Gaussian Splatting}, 
      author={Jiawei Ren and Liang Pan and Jiaxiang Tang and Chi Zhang and Ang Cao and Gang Zeng and Ziwei Liu},
      year={2024},
      eprint={2312.17142},
      archivePrefix={arXiv},
      primaryClass={cs.CV},
      url={https://arxiv.org/abs/2312.17142}, 
}

@article{fisher2021colmap,
  title={ColMap: A memory-efficient occupancy grid mapping framework},
  author={Fisher, Alex and Cannizzaro, Ricardo and Cochrane, Madeleine and Nagahawatte, Chatura and Palmer, Jennifer L},
  journal={Robotics and Autonomous Systems},
  volume={142},
  pages={103755},
  year={2021},
  publisher={Elsevier}
}

@article{orbslam,
  title={ORB-SLAM: A versatile and accurate monocular SLAM system},
  author={Mur-Artal, Raul and Montiel, Jose Maria Martinez and Tardos, Juan D},
  journal={IEEE transactions on robotics},
  volume={31},
  pages={1147--1163},
  year={2015},
  publisher={IEEE}
}

@inproceedings{4dgs,
  title={4d gaussian splatting for real-time dynamic scene rendering},
  author={Wu, Guanjun and Yi, Taoran and Fang, Jiemin and Xie, Lingxi and Zhang, Xiaopeng and Wei, Wei and Liu, Wenyu and Tian, Qi and Wang, Xinggang},
  booktitle={Proceedings of the IEEE/CVF conference on computer vision and pattern recognition},
  pages={20310--20320},
  year={2024}
}

@article{jiao2026large,
  title={Large-kernel spatially parallel feature fusion for monocular 3D perception in autonomous driving},
  author={Jiao, Ruanzhi and Zhang, Jinlai and Li, Chang and Hu, Lin},
  journal={Knowledge-Based Systems},
  volume={343},
  pages={115998},
  year={2026},
  publisher={Elsevier}
}

@article{zhang2026multivariate,
  title={Multivariate feature learning and associative spatial information enhancement for snow object detection in autonomous driving},
  author={Zhang, Jinlai and Xiang, Mingchao and Hu, Yongheng and Hao, Wei and Lei, Linlong and Yi, Kefu},
  journal={Engineering Applications of Artificial Intelligence},
  volume={175},
  pages={114672},
  year={2026},
  publisher={Elsevier}
}

@article{zhang2026adaptive,
  title={Adaptive dual cross-attention network for multispectral object detection in autonomous driving},
  author={Zhang, Jinlai and Song, Xiaolong and Li, Yucheng and Liang, Diqing and Zhang, Zhiyong and Cai, Jinhu},
  journal={Expert Systems with Applications},
  pages={132012},
  year={2026},
  publisher={Elsevier}
}

@article{hu2026anyslot,
  title={AnySlot: Goal-Conditioned Vision-Language-Action Policies for Zero-Shot Slot-Level Placement},
  author={Hu, Zhaofeng and Zhou, Sifan and Zhang, Qinbo and Xu, Rongtao and Su, Qi and Liang, Ci-Jyun},
  journal={arXiv preprint arXiv:2604.10432},
  year={2026}
}

@inproceedings{zhao2025tartan,
  title={Tartan imu: A light foundation model for inertial positioning in robotics},
  author={Zhao, Shibo and Zhou, Sifan and Blanchard, Raphael and Qiu, Yuheng and Wang, Wenshan and Scherer, Sebastian},
  booktitle={Proceedings of the Computer Vision and Pattern Recognition Conference},
  pages={22520--22529},
  year={2025}
}

@article{zhao2025resilient,
  title={Resilient odometry via hierarchical adaptation},
  author={Zhao, Shibo and Zhou, Sifan and Zhang, Yuchen and Zhang, Ji and Wang, Chen and Wang, Wenshan and Scherer, Sebastian},
  journal={Science Robotics},
  volume={10},
  number={109},
  pages={eadv1818},
  year={2025},
  publisher={American Association for the Advancement of Science}
}

@article{zhao2026advances,
  title={Advances in Global Solvers for 3D Vision},
  author={Zhao, Zhenjun and Yang, Heng and Liao, Bangyan and Zeng, Yingping and Yan, Shaocheng and Gu, Yingdong and Liu, Peidong and Zhou, Yi and Li, Haoang and Civera, Javier},
  journal={arXiv preprint arXiv:2602.14662},
  year={2026}
}

@inproceedings{zhao2024balf,
  title={Balf: Simple and efficient blur aware local feature detector},
  author={Zhao, Zhenjun},
  booktitle={Proceedings of the IEEE/CVF Winter Conference on Applications of Computer Vision},
  pages={3362--3372},
  year={2024}
}

@inproceedings{zhao2023benchmark,
  title={Benchmark for Evaluating Initialization of Visual-Inertial Odometry},
  author={Zhao, Zhenjun and Chen, Ben M},
  booktitle={2023 42nd Chinese Control Conference (CCC)},
  pages={3935--3940},
  year={2023},
  organization={IEEE}
}

@article{zeng2025janusvln,
            title={JanusVLN: Decoupling Semantics and Spatiality with Dual Implicit Memory for Vision-Language Navigation},
            author={Zeng, Shuang and Qi, Dekang and Chang, Xinyuan and Xiong, Feng and Xie, Shichao and Wu, Xiaolong and Liang, Shiyi and Xu, Mu and Wei, Xing},
            journal={arXiv preprint arXiv:2509.22548},
            year={2025}
            }

@article{zeng2025FSDrive,
      title={FutureSightDrive: Thinking Visually with Spatio-Temporal CoT for Autonomous Driving},
      author={Shuang Zeng and Xinyuan Chang and Mengwei Xie and Xinran Liu and Yifan Bai and Zheng Pan and Mu Xu and Xing Wei},
      journal={arXiv preprint arXiv:2505.17685},
      year={2025}
      }

@inproceedings{wang2025vggt,
  title={Vggt: Visual geometry grounded transformer},
  author={Wang, Jianyuan and Chen, Minghao and Karaev, Nikita and Vedaldi, Andrea and Rupprecht, Christian and Novotny, David},
  booktitle={Proceedings of the Computer Vision and Pattern Recognition Conference},
  pages={5294--5306},
  year={2025}
}

@inproceedings{chenttt3r,
  title={TTT3R: 3D Reconstruction as Test-Time Training},
  author={Chen, Xingyu and Chen, Yue and Xiu, Yuliang and Geiger, Andreas and Chen, Anpei},
  booktitle={The Fourteenth International Conference on Learning Representations}
}

@article{yuan2025test3r,
  title={Test3r: Learning to reconstruct 3d at test time},
  author={Yuan, Yuheng and Shen, Qiuhong and Wang, Shizun and Yang, Xingyi and Wang, Xinchao},
  journal={arXiv preprint arXiv:2506.13750},
  volume={5},
  year={2025}
}

@misc{xiong2025autoregressivemodelsvisionsurvey,
      title={Autoregressive Models in Vision: A Survey}, 
      author={Jing Xiong and Gongye Liu and Lun Huang and Chengyue Wu and Taiqiang Wu and Yao Mu and Yuan Yao and Hui Shen and Zhongwei Wan and Jinfa Huang and Chaofan Tao and Shen Yan and Huaxiu Yao and Lingpeng Kong and Hongxia Yang and Mi Zhang and Guillermo Sapiro and Jiebo Luo and Ping Luo and Ngai Wong},
      year={2025},
      eprint={2411.05902},
      archivePrefix={arXiv},
      primaryClass={cs.CV},
      url={https://arxiv.org/abs/2411.05902}, 
}

@inproceedings{van2016pixel,
  title={Pixel recurrent neural networks},
  author={Van Den Oord, A{\"a}ron and Kalchbrenner, Nal and Kavukcuoglu, Koray},
  booktitle={International conference on machine learning},
  pages={1747--1756},
  year={2016},
  organization={PMLR}
}

@misc{villegas2022phenakivariablelengthvideo,
      title={Phenaki: Variable Length Video Generation From Open Domain Textual Description}, 
      author={Ruben Villegas and Mohammad Babaeizadeh and Pieter-Jan Kindermans and Hernan Moraldo and Han Zhang and Mohammad Taghi Saffar and Santiago Castro and Julius Kunze and Dumitru Erhan},
      year={2022},
      eprint={2210.02399},
      archivePrefix={arXiv},
      primaryClass={cs.CV},
      url={https://arxiv.org/abs/2210.02399}, 
}

@article{VAR,
  title={Visual autoregressive modeling: Scalable image generation via next-scale prediction},
  author={Tian, Keyu and Jiang, Yi and Yuan, Zehuan and Peng, Bingyue and Wang, Liwei},
  journal={Advances in neural information processing systems},
  volume={37},
  pages={84839--84865},
  year={2024}
}

@inproceedings{henschel2025streamingt2v,
  title={Streamingt2v: Consistent, dynamic, and extendable long video generation from text},
  author={Henschel, Roberto and Khachatryan, Levon and Poghosyan, Hayk and Hayrapetyan, Daniil and Tadevosyan, Vahram and Wang, Zhangyang and Navasardyan, Shant and Shi, Humphrey},
  booktitle={Proceedings of the Computer Vision and Pattern Recognition Conference},
  pages={2568--2577},
  year={2025}
}

@article{wu2025point3r,
  title={Point3r: Streaming 3d reconstruction with explicit spatial pointer memory},
  author={Wu, Yuqi and Zheng, Wenzhao and Zhou, Jie and Lu, Jiwen},
  journal={arXiv preprint arXiv:2507.02863},
  year={2025}
}

@inproceedings{sandstrom2023point,
  title={Point-slam: Dense neural point cloud-based slam},
  author={Sandstr{\"o}m, Erik and Li, Yue and Van Gool, Luc and Oswald, Martin R},
  booktitle={Proceedings of the IEEE/CVF international conference on computer vision},
  pages={18433--18444},
  year={2023}
}

@inproceedings{cheng2024gaussianpro,
  title={Gaussianpro: 3d gaussian splatting with progressive propagation},
  author={Cheng, Kai and Long, Xiaoxiao and Yang, Kaizhi and Yao, Yao and Yin, Wei and Ma, Yuexin and Wang, Wenping and Chen, Xuejin},
  booktitle={Forty-first International Conference on Machine Learning},
  year={2024}
}

@article{kerbl20233d,
  title={3D Gaussian splatting for real-time radiance field rendering.},
  author={Kerbl, Bernhard and Kopanas, Georgios and Leimk{\"u}hler, Thomas and Drettakis, George},
  journal={ACM Trans. Graph.},
  volume={42},
  number={4},
  pages={139--1},
  year={2023}
}

@article{yin20234dgen,
  title={4dgen: Grounded 4d content generation with spatial-temporal consistency},
  author={Yin, Yuyang and Xu, Dejia and Wang, Zhangyang and Zhao, Yao and Wei, Yunchao},
  journal={arXiv preprint arXiv:2312.17225},
  year={2023}
}

@inproceedings{wu2025cat4d,
  title={Cat4d: Create anything in 4d with multi-view video diffusion models},
  author={Wu, Rundi and Gao, Ruiqi and Poole, Ben and Trevithick, Alex and Zheng, Changxi and Barron, Jonathan T and Holynski, Aleksander},
  booktitle={Proceedings of the Computer Vision and Pattern Recognition Conference},
  pages={26057--26068},
  year={2025}
}

@misc{huang2025selfforcingbridgingtraintest,
      title={Self Forcing: Bridging the Train-Test Gap in Autoregressive Video Diffusion}, 
      author={Xun Huang and Zhengqi Li and Guande He and Mingyuan Zhou and Eli Shechtman},
      year={2025},
      eprint={2506.08009},
      archivePrefix={arXiv},
      primaryClass={cs.CV},
      url={https://arxiv.org/abs/2506.08009}, 
}

@misc{yin2025slowbidirectionalfastautoregressive,
      title={From Slow Bidirectional to Fast Autoregressive Video Diffusion Models}, 
      author={Tianwei Yin and Qiang Zhang and Richard Zhang and William T. Freeman and Fredo Durand and Eli Shechtman and Xun Huang},
      year={2025},
      eprint={2412.07772},
      archivePrefix={arXiv},
      primaryClass={cs.CV},
      url={https://arxiv.org/abs/2412.07772}, 
}

@misc{liu2025rollingforcingautoregressivelong,
      title={Rolling Forcing: Autoregressive Long Video Diffusion in Real Time}, 
      author={Kunhao Liu and Wenbo Hu and Jiale Xu and Ying Shan and Shijian Lu},
      year={2025},
      eprint={2509.25161},
      archivePrefix={arXiv},
      primaryClass={cs.CV},
      url={https://arxiv.org/abs/2509.25161}, 
}

@article{gu2025starflow,
  title={STARFlow-V: End-to-End Video Generative Modeling with Normalizing Flows},
  author={Gu, Jiatao and Shen, Ying and Chen, Tianrong and Dinh, Laurent and Wang, Yuyang and Bautista, Miguel Angel and Berthelot, David and Susskind, Josh and Zhai, Shuangfei},
  journal={arXiv preprint arXiv:2511.20462},
  year={2025}
}

@article{zhuo2025streaming,
  title={Streaming 4d visual geometry transformer},
  author={Zhuo, Dong and Zheng, Wenzhao and Guo, Jiahe and Wu, Yuqi and Zhou, Jie and Lu, Jiwen},
  journal={arXiv preprint arXiv:2507.11539},
  year={2025}
}

@article{chen2025teleworld,
  title={TeleWorld: Towards Dynamic Multimodal Synthesis with a 4D World Model},
  author={Chen, Yabo and Liang, Yuanzhi and Wang, Jiepeng and Chen, Tingxi and Cheng, Junfei and Gu, Zixiao and Huang, Yuyang and Jiang, Zicheng and Li, Wei and Li, Tian and others},
  journal={arXiv preprint arXiv:2601.00051},
  year={2025}
}

@misc{cheng2025playingtransformer30fps,
      title={Playing with Transformer at 30+ FPS via Next-Frame Diffusion}, 
      author={Xinle Cheng and Tianyu He and Jiayi Xu and Junliang Guo and Di He and Jiang Bian},
      year={2025},
      eprint={2506.01380},
      archivePrefix={arXiv},
      primaryClass={cs.CV},
      url={https://arxiv.org/abs/2506.01380}, 
}

@article{zhang2024monst3r,
  title={Monst3r: A simple approach for estimating geometry in the presence of motion},
  author={Zhang, Junyi and Herrmann, Charles and Hur, Junhwa and Jampani, Varun and Darrell, Trevor and Cole, Forrester and Sun, Deqing and Yang, Ming-Hsuan},
  journal={arXiv preprint arXiv:2410.03825},
  year={2024}
}

@inproceedings{Spann3R,
  title={3d reconstruction with spatial memory},
  author={Wang, Hengyi and Agapito, Lourdes},
  booktitle={2025 International Conference on 3D Vision (3DV)},
  pages={78--89},
  year={2025},
  organization={IEEE}
}

@inproceedings{7-scenes,
  title={Scene coordinate regression forests for camera relocalization in RGB-D images},
  author={Shotton, Jamie and Glocker, Ben and Zach, Christopher and Izadi, Shahram and Criminisi, Antonio and Fitzgibbon, Andrew},
  booktitle={Proceedings of the IEEE conference on computer vision and pattern recognition},
  pages={2930--2937},
  year={2013}
}

@inproceedings{wang2025continuous,
  title={Continuous 3d perception model with persistent state},
  author={Wang, Qianqian and Zhang, Yifei and Holynski, Aleksander and Efros, Alexei A and Kanazawa, Angjoo},
  booktitle={Proceedings of the Computer Vision and Pattern Recognition Conference},
  pages={10510--10522},
  year={2025}
}

@article{zhu2025ar4d,
  title={Ar4d: Autoregressive 4d generation from monocular videos},
  author={Zhu, Hanxin and He, Tianyu and Yu, Xiqian and Guo, Junliang and Chen, Zhibo and Bian, Jiang},
  journal={arXiv preprint arXiv:2501.01722},
  year={2025}
}

@article{chen20254dnex,
  title={4dnex: Feed-forward 4d generative modeling made easy},
  author={Chen, Zhaoxi and Liu, Tianqi and Zhuo, Long and Ren, Jiawei and Tao, Zeng and Zhu, He and Hong, Fangzhou and Pan, Liang and Liu, Ziwei},
  journal={arXiv preprint arXiv:2508.13154},
  year={2025}
}

@article{wang20254real,
  title={4Real-Video-V2: Fused View-Time Attention and Feedforward Reconstruction for 4D Scene Generation},
  author={Wang, Chaoyang and Mirzaei, Ashkan and Goel, Vidit and Menapace, Willi and Siarohin, Aliaksandr and Vinella, Avalon and Vasilkovsky, Michael and Skorokhodov, Ivan and Shakhrai, Vladislav and Korolev, Sergey and others},
  journal={arXiv preprint arXiv:2506.18839},
  year={2025}
}

@article{pan2025diff4splat,
  title={Diff4Splat: Controllable 4D Scene Generation with Latent Dynamic Reconstruction Models},
  author={Pan, Panwang and Lin, Chenguo and Zhao, Jingjing and Li, Chenxin and Lin, Yuchen and Li, Haopeng and Yan, Honglei and Wen, Kairun and Lin, Yunlong and Yuan, Yixuan and others},
  journal={arXiv preprint arXiv:2511.00503},
  year={2025}
}

@article{liu2025dream4d,
  title={Dream4D: Lifting Camera-Controlled I2V towards Spatiotemporally Consistent 4D Generation},
  author={Liu, Xiaoyan and Li, Kangrui and Song, Yuehao and Liu, Jiaxin},
  journal={arXiv preprint arXiv:2508.07769},
  year={2025}
}
}
\clearpage
\setcounter{page}{1}
% \maketitlesupplementary

\twocolumn[{
    \centering
    {\Large \textbf{Streaming4D: Accelerate 4D World Models via Block-wise Video Generation and Incremental Reconstruction}} \\[0.8em]
    {\large Supplementary Material} \\[1.5em]
    
    Xiaoyan Liu$^{1\dagger}$, 
    Kangrui Li$^{2\dagger}$, 
    Jiaxin Liu$^{3\dagger}$, 
    Sifan Zhou$^{4*}$ \\[0.6em]
    
    $^1$The Chinese University of Hong Kong \quad
    $^2$The Hong Kong Polytechnic University \\[0.3em]
    $^3$The University of New South Wales \quad
    $^4$Southeast University \\[0.8em]
    
    {\small $\dagger$ Equal contribution. \quad $*$ Corresponding author.} \\[0.5em]
    
    {\footnotesize 
    \texttt{liuxy185@link.cuhk.edu.hk}, 
    \texttt{24120659G@connect.polyu.hk}, \\
    \texttt{z5565763@ad.unsw.edu.au}, 
    \texttt{sifanjay@gmail.com}}
    
    \vspace{1.5em}  % 与正文之间的间距
}]

\appendix
\section{Latency Optimization and Hardware Contention.}
\label{sec:LOHC}

As mentioned in \cref{sec:3.3}, in an ideal decoupled environment, the cumulative latency for a sequence of $K$ blocks would be optimized from a sequential sum $\sum \left( \mathcal{T}_{\text{gen}}(B_i) + \mathcal{T}_{\text{rec}}(B_i) \right)$ to a parallel execution time bounded by $\sum \max\big(\mathcal{T}_{\text{gen}}, \mathcal{T}_{\text{rec}})$. However, real-time 4D synthesis is typically deployed on a single GPU under strict hardware constraints.

When executing the generative module $G_{video}$ and the reconstruction module $R_{4D}$ concurrently through asynchronous execution, the two heavily parameterized models inevitably compete for limited Streaming Multiprocessors (SMs) and memory bandwidth. Let $\alpha > 0$ denote the performance penalty induced by this contention of resources. The practical execution latency $\tilde{\mathcal{L}}$ is modeled as:

\begin{equation}
\begin{split}
\tilde{\mathcal{L}}_{\text{ours}} \approx{} &
\mathcal{T}_{\text{gen}}(B_1) + \sum_{i=2}^{K} \max \left( \tilde{\mathcal{T}}_{\text{gen}}(B_i), \tilde{\mathcal{T}}_{\text{rec}}(B_{i-1}) \right) \\
& + \tilde{\mathcal{T}}_{\text{rec}}(B_K),
\end{split}
\end{equation}

where $\tilde{\mathcal{L}} = (1+\alpha) \mathcal{L}$ represents the inflated execution time under concurrent workload. 

Despite the presence of hardware contention $\alpha$, our pipelined strategy still significantly outperforms the sequential baseline. By overlapping the generative denoising with the intensive geometric reasoning, our system effectively hides a substantial portion of the video generation overhead. 

\section{Analysis of Pipeline Efficiency across Resolutions.}
\label{sec:APE}

\cref{tab:time_comparison} presents the generation time and speedup ratios of our proposed pipeline compared to the sequential baseline across different input resolutions on a single RTX 4090 GPU. 
Interestingly, we observe an inverted-U performance curve in relative speedup, which peaks at $1.24\times$ at the $512 \times 288$ resolution, before plateauing or slightly degrading at both lower and higher extremes.

This phenomenon perfectly reflects the trade-off between pipeline overlap and hardware resource contention (as theoretically modeled by $\alpha$ in \cref{sec:LOHC}.

At lower resolutions (e.g., $384\times208$), the absolute computational workload is relatively light. Consequently, constant system-level overheads occupy a larger proportion of the execution time, thereby diluting the benefits of parallel execution. In contrast, at higher resolutions, the quadratic complexity of the Transformer-based generator and the 3D reconstructor aggressively saturates the GPU's memory bandwidth and Streaming Multiprocessors (SMs). This severe resource contention inflates the individual execution time of each module, bounding the theoretical maximum speedup. Consequently, the optimal efficiency is observed at certain moderate resolutions, where the workload is substantial enough to hide scheduling overheads, yet still fits comfortably within the compute and memory limits of a single GPU, thereby maximizing pipeline overlap.

Nevertheless, our pipeline consistently maintains a robust $\sim 20\%$ overall speedup across all tested resolutions, validating its resilience and scalability in single-GPU interactive 4D environments.

\section{Architecture}
\label{archi}
We employ an integrated tightly coupled synchronous streaming pipeline to enable real-time synthesis of 4D content. This architecture consists of two core modules. At the front end of the architecture, we employ the Self Forcing~\cite{huang2025selfforcingbridgingtraintest} paradigm for generating video blocks. Instead of generating the entire video sequence in a single pass, the model continuously synthesizes frames in a streaming manner and groups every three consecutive frames into a temporal block. Each block functions as a compact spatiotemporal unit that captures local motion and appearance dynamics while enabling seamless integration with the subsequent 4D reconstruction stage. 
% maintaining causal dependencies across blocks through cached key–value states. 
% This design preserves the efficiency of the Self-forcing rollout and enables seamless integration with the subsequent 4D reconstruction stage.

% Building upon the Self-forcing architecture~\cite{huang2025selfforcingbridgingtraintest}, we develop a streaming block-wise autoregressive framework for real-time 4D synthesis. Instead of generating the entire video sequence in a single pass, the model continuously synthesizes frames in a streaming manner and groups every three consecutive frames into a temporal block. Each block functions as a compact spatiotemporal unit that captures local motion and appearance dynamics while maintaining causal dependencies across blocks through cached key–value states. This design preserves the efficiency of the Self-forcing rollout and enables seamless integration with the subsequent 4D reconstruction stage.

In the backend, we incorporate a module based on the CUT3R~\cite{wang2025continuous} architecture for incremental 4D scene update. For each generated block, the 4D generator incrementally updates a persistent world state that encodes evolving geometry, depth, and appearance features, producing a temporally coherent 4D representation. Through the integration of AR video generation and 4D reconstruction, our framework delivers real-time, geometrically consistent, and temporally stable 4D content generation, outperforming conventional autoregressive or reconstruction-only methods in both efficiency and long-term coherence.

\section{Future Work.}
\label{sec:FW}
Although our current framework successfully establishes a forward-streaming pipeline from autoregressive video generation (Self Forcing) to 3D scene reconstruction (CUT3R), we envision a future trajectory that fully activates the geometric guidance path (as indicated in \cref{fig:pipline}). Relying solely on 2D pixel-space history, long-duration autoregressive video generators are inherently susceptible to long-term geometric drift and structural hallucinations. 

To further refine geometric consistency, one could incorporate test-time optimization strategies, similar to Test3R~\cite{yuan2025test3r} or TTT3R~\cite{chenttt3r}, which adapt the model parameters online to minimize reprojection errors for a specific input stream. Building upon this improved per-frame geometry, the system can then leverage the refined 3D information for long-term coherence. To further enhance the generation and reconstruction quality over extended durations, we plan to feed the continuously updated 3D scene information from our existing persistent world memory back into the video generation stage. Serving as an explicit geometric prior, this feedback will guide the autoregressive model to evolve from merely predicting local 2D fragments to simulating a coherent, explorable dynamic world. Ultimately, we anticipate that this cyclic guidance mechanism will effectively mitigate visual drift and ensure robust spatiotemporal coherence across infinite-length sequences.

% As demonstrated in our experiments, this architectural co-design yields a tangible net reduction in end-to-end latency, making the otherwise prohibitive 4D autoregressive generation viable for interactive applications on a single GPU.

% Having the supplementary compiled together with the main paper means that:
% % 
% \begin{itemize}
% \item The supplementary can back-reference sections of the main paper, for example, we can refer to \cref{sec:intro};
% \item The main paper can forward reference sub-sections within the supplementary explicitly (e.g. referring to a particular experiment); 
% \item When submitted to arXiv, the supplementary will already included at the end of the paper.
% \end{itemize}
% % 
% To split the supplementary pages from the main paper, you can use \href{https://support.apple.com/en-ca/guide/preview/prvw11793/mac#:~:text=Delete%20a%20page%20from%20a,or%20choose%20Edit%20%3E%20Delete).}{Preview (on macOS)}, \href{https://www.adobe.com/acrobat/how-to/delete-pages-from-pdf.html#:~:text=Choose%20%E2%80%9CTools%E2%80%9D%20%3E%20%E2%80%9COrganize,or%20pages%20from%20the%20file.}{Adobe Acrobat} (on all OSs), as well as \href{https://superuser.com/questions/517986/is-it-possible-to-delete-some-pages-of-a-pdf-document}{command line tools}.

% WARNING: do not forget to delete the supplementary pages from your submission 
% \input{sec/X_suppl}

\end{document}